\documentclass[sigconf]{acmart}

\AtBeginDocument{%
  }

\setcopyright{rightsretained}
\copyrightyear{2024}
\acmYear{2024}
\begin{document}

\title[LOttery Transformers with Ultra Sparsity]{LOTUS: Improving Transformer Efficiency with Sparsity Pruning  and Data Lottery Tickets}

\author{Ojasw Upadhyay}
\email{ojasw@gatech.edu}
\affiliation{%
  \institution{Georgia Institute of Technology}
  \city{Atlanta}
  \state{Georgia}
  \country{USA}
}

\renewcommand{\shortauthors}{Upadhyay, Ojasw}

\begin{abstract}
  Vision transformers have revolutionized computer vision, but their computational demands present challenges for training and deployment. This paper introduces LOTUS (LOttery Transformers with Ultra Sparsity), a novel method that leverages data lottery ticket selection and sparsity pruning to accelerate vision transformer training while maintaining accuracy. Our approach focuses on identifying and utilizing the most informative data subsets and eliminating redundant model parameters to optimize the training process. Through extensive experiments, we demonstrate the effectiveness of LOTUS in achieving rapid convergence and high accuracy with significantly reduced computational requirements. This work highlights the potential of combining data selection and sparsity techniques for efficient vision transformer training, opening doors for further research and development in this area.
\end{abstract}


\begin{teaserfigure}
  \includegraphics[width=\textwidth]{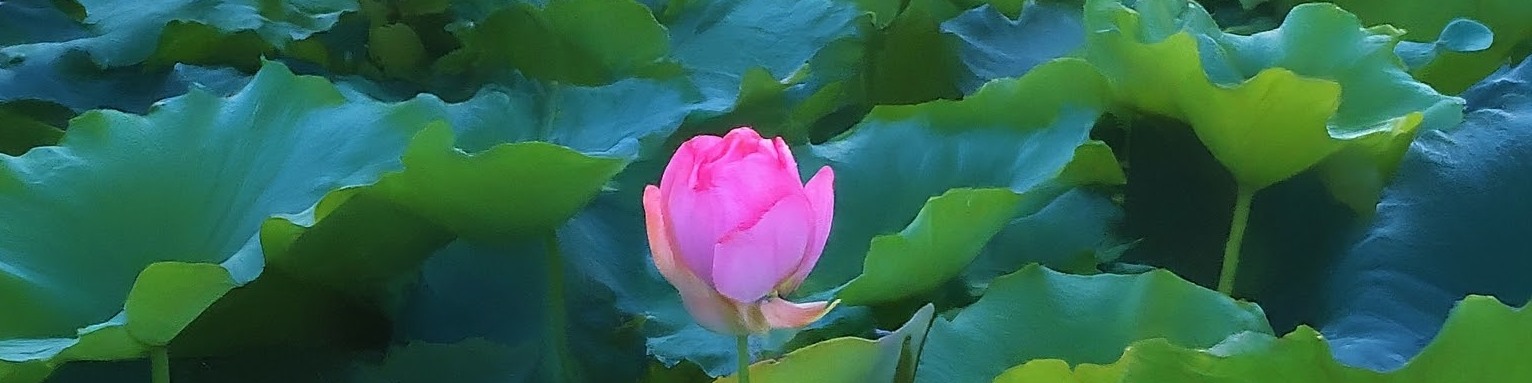}
  \caption{A sparse wetland with a single lotus flower.}
  \label{fig:teaser}
\end{teaserfigure}

\received{1 May 2024}

\maketitle

\section{Introduction}
Deep learning, particularly transformers \cite{Vaswani2017}, revolutionized natural language processing (NLP) and vision tasks. In particular, Vision transformers have emerged as powerful tools in computer vision, demonstrating impressive performance across various tasks like image classification and object detection. However, their eﬀectiveness is often accompanied by significant computational demands due to their large size and complex architectures. This presents a challenge, especially when dealing with limited resources or time constraints. With the increasing popularity of transformers in vision tasks, the need for efficient training methods for vision transformers has become crucial. This research aims to address this challenge by developing novel techniques to improve the efficiency of vision transformer training, focusing on reducing training time while maintaining or enhancing performance.

\section{Existing Works and Limitations}
Several approaches address training efficiency, including Instant Soup Pruning and Essential Sparsity. Instant Soup Pruning (ISP) \cite{ISP2023} explores the idea of merging multiple weak models to create a strong, sparse model, eﬀectively reducing computational requirements. Essential Sparsity \cite{EssentialSparsity2023} investigates the inherent sparsity within large pre-trained models, revealing that a significant portion of weights can be removed without impacting performance. While these techniques show promise, they have primarily focused on non-transformer architectures. This research aims to bridge this gap by applying and adapting these methods to vision transformers, exploring their synergy for efficient training.

In addition to pruning techniques, Dynamic-ViT and Data Lottery Tickets are promising techniques. Dynamic-ViT \cite{DynamicViT2021} introduces a dynamic token reduction mechanism that adaptively reduces the number of tokens in the input sequence, improving the efficiency of transformer models. Data Lottery Tickets \cite{DataLevelLottery2023} suggest the existence of smaller subsets of training data that are sufficient for achieving comparable accuracy to the full dataset. Combining token reduction with the data lottery ticket concept could further enhance the efficiency of transformer training. This research will explore the potential of combining these techniques to develop a novel approach for efficient transformer training.

\section{Methods}
We propose LOTUS (LOttery Transformers with Ultra Sparsity), a novel approach that leverages pruning techniques and data lottery tickets to improve the efficiency of transformer training. There are three main components of the LOTUS approach.

The first phase involves identifying data lottery tickets using attention maps to determine the most informative data patches for training. By selecting a subset of data patches that are crucial for training, we aim to reduce the training time while maintaining accuracy levels because the model focuses on the most relevant information as it trains as opposed to the entire dataset.

The second phase involves applying Instant Sparse Soup Pruning (ISSP) to the transformer model based on the identified data lottery tickets. We use magnitude-based pruning to remove unimportant weights from the model, reducing the number of parameters and computational requirements. The first round of pruning uses the essential sparsity number found through pruning the pretrained model at various sparsity levels. The second round of pruning uses Instant Soup Pruning (ISP) to further reduce the model size. To create the denoised mask for ISP, the model is trained on 10\% of the data and the prune mask is created based on the weights that are not pruned in the first round. The denoised mask is the merge of the masks created at different sparsity levels using the union of the masks. This two-step pruning process aims to achieve significant sparsity while theoretically maintaining similar performance.

The third phase involves fine-tuning the pruned model using the remaining data patches. We aim to demonstrate that the pruned model can achieve comparable or even better performance than the baseline model while significantly reducing the number of epochs required for training.

\section{Results}
\begin{figure}[h]
  \centering
  \includegraphics[width=0.9\linewidth]{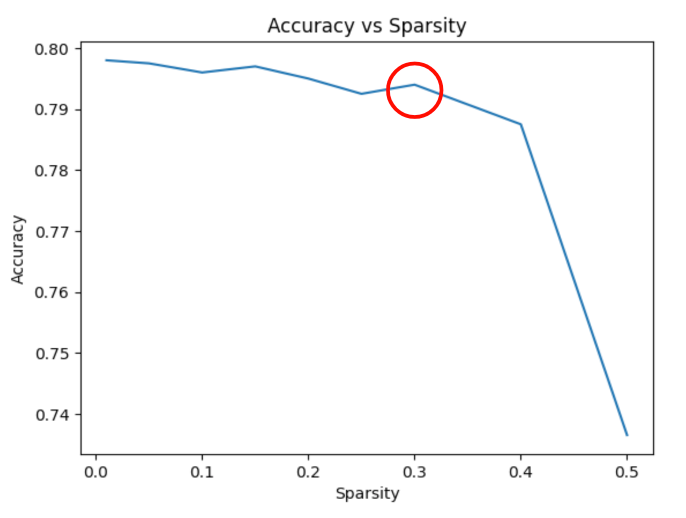}
  \caption{The plot shows the accuracy of the model at different sparsity levels.}
  \label{fig:essential-sparsity}
\end{figure}

From the experiments conducted, we started with a pre-trained ViT model on the CIFAR-10 dataset. We applied one-shot magnitude pruning to the model at various sparsity levels ranging from 1\% to 50\%. The model pruned at 30\% sparsity maintained a high accuracy of around 79\% (minimal degradation compared to the baseline). We can see in the graph \ref{fig:essential-sparsity} that 30\% sparsity is the optimal sparsity level for the model without significant loss in accuracy.

\begin{figure}[h]
    \centering
    \includegraphics[width=0.9\linewidth]{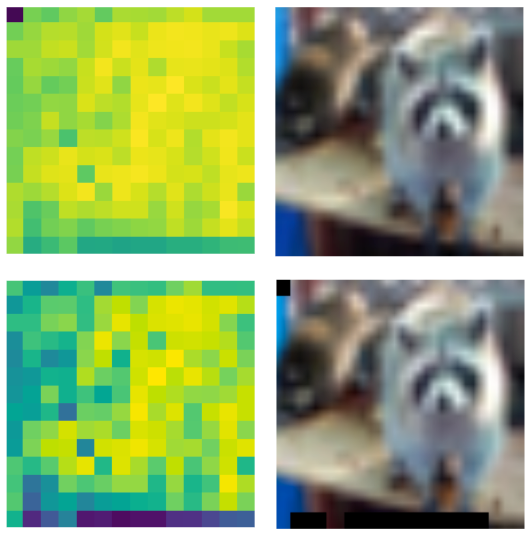}
    \caption{An example of data lottery tickets created using attention maps with 10\% of the data patches removed.}
    \label{fig:lottery-tickets}
\end{figure}

In the second phase, we successfully created data lottery tickets using attention maps from the pre-trained model. The lottery data was created by removing X\% of the data patches that had the lowest attention scores. The example in figure \ref{fig:lottery-tickets} shows an example with 10\% of the data patches removed. The first token of the attention maps had to be normalized to the mean to avoid the attention sink, making the distribution of the attention scores more uniform.

\begin{figure}[h]
    \centering
    \includegraphics[width=0.9\linewidth]{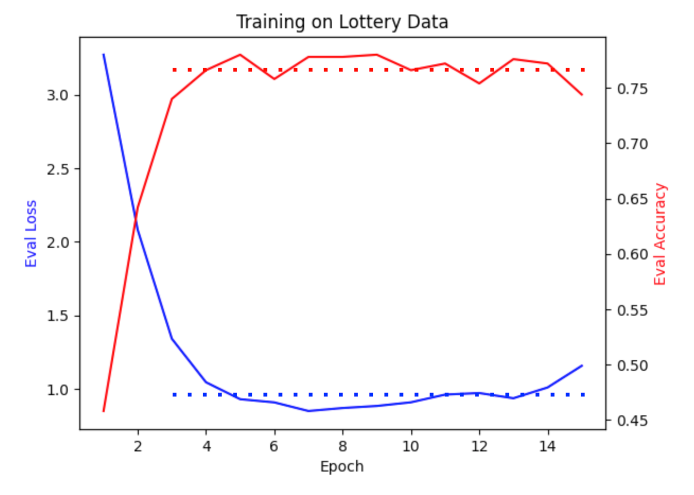}
    \caption{The accuracy and loss plots of the model fine-tuned on the data-level lottery tickets.}
    \label{fig:lottery-training}
\end{figure}

Then, the untrained model was run on the data-level lottery tickets for many epochs using the remaining data patches. The fine-tuning process showed rapid convergence, with the model achieving close to SOTA performance by only the fifth epoch. The accuracy and loss plots in figure \ref{fig:lottery-training} show the model's rapid convergence and high accuracy after fine-tuning on the lottery data.

\begin{figure}[h]
    \centering
    \includegraphics[width=0.9\linewidth]{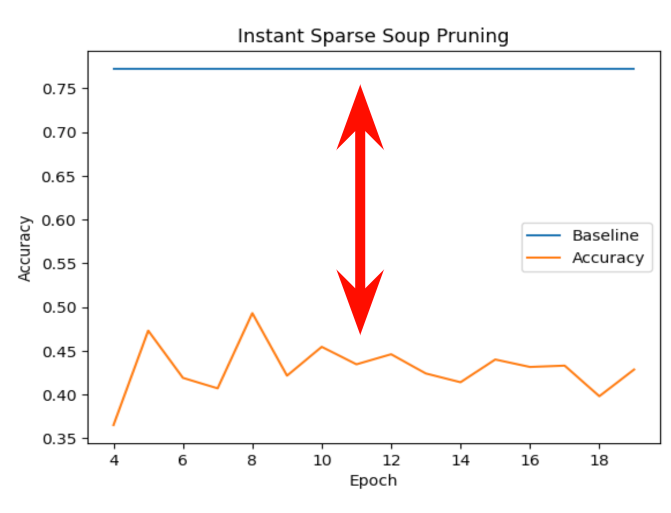}
    \caption{The accuracy of the model after applying the ISSP approach.}
    \label{fig:issp}
\end{figure}

However, the ISSP approach led to the model's accuracy being significantly lower compared to the baseline. The model's accuracy was around 50\% after both one-shot Essential Sparsity pruning and Instant Soup Pruning, which is a significant drop compared to the baseline model as shown in figure \ref{fig:issp}. The reasoning behind this drop in accuracy could be due to the aggressive pruning of the model, which had started to remove more than 50\% of the weights. This aggressive pruning would have removed important weights that were crucial for the model's performance, leading to the drop in accuracy. Moreover, the denoised mask created for the ISP approach might not have been optimal as the model was trained using only 10\% of the data and the model might not have started converging towards the optimal solution. More experiments need to be conducted to optimize the pruning process and the denoised mask creation to achieve better results with the ISSP approach.

\section{Conclusion}
While data lottery tickets shows rapid convergence with a pruned model, the ISSP approach did not perform and further investigation is needed to determine what is contributing to the performance discrepancy and to explore modifications or alternative strategies. This research highlights the promise of data lottery tickets for efficient training of vision transformers, paving the way for further exploration and optimization of this approach.

% \section{Tables}

% Immediately following this sentence is the point at which
% Table~\ref{tab:freq} is included in the input file; compare the
% placement of the table here with the table in the printed output of
% this document.

% \begin{table}
%   \caption{Frequency of Special Characters}
%   \label{tab:freq}
%   \begin{tabular}{ccl}
%     \toprule
%     Non-English or Math&Frequency&Comments\\
%     \midrule
%     \O & 1 in 1,000& For Swedish names\\
%     $\pi$ & 1 in 5& Common in math\\
%     \$ & 4 in 5 & Used in business\\
%     $\Psi^2_1$ & 1 in 40,000& Unexplained usage\\
%   \bottomrule
% \end{tabular}
% \end{table}

% To set a wider table,

% \begin{table*}
%   \caption{Some Typical Commands}
%   \label{tab:commands}
%   \begin{tabular}{ccl}
%     \toprule
%     Command &A Number & Comments\\
%     \midrule
%     \texttt{{\char'134}author} & 100& Author \\
%     \texttt{{\char'134}table}& 300 & For tables\\
%     \texttt{{\char'134}table*}& 400& For wider tables\\
%     \bottomrule
%   \end{tabular}
% \end{table*}

\begin{acks}
To Celine Lin and the TAs, for their class on efficient machine learning that inspired this research. To Georgia Tech for providing the resources for this research.
\end{acks}

\bibliographystyle{ACM-Reference-Format}
\bibliography{ojasw-eml-paper}


\begin{thebibliography}{5}

%%% the Web address.

\ifx \showCODEN    \undefined \def \showCODEN     #1{\unskip}     \fi
\ifx \showDOI      \undefined \def \showDOI       #1{#1}\fi
\ifx \showISBNx    \undefined \def \showISBNx     #1{\unskip}     \fi
\ifx \showISBNxiii \undefined \def \showISBNxiii  #1{\unskip}     \fi
\ifx \showISSN     \undefined \def \showISSN      #1{\unskip}     \fi
\ifx \showLCCN     \undefined \def \showLCCN      #1{\unskip}     \fi
\ifx \shownote     \undefined \def \shownote      #1{#1}          \fi
\ifx \showarticletitle \undefined \def \showarticletitle #1{#1}   \fi
\ifx \showURL      \undefined \def \showURL       {\relax}        \fi
% invisible to TeX
\providecommand\bibfield[2]{#2}
\providecommand\bibinfo[2]{#2}
\providecommand\natexlab[1]{#1}
\providecommand\showeprint[2][]{arXiv:#2}

\bibitem[Jaiswal et~al\mbox{.}(2023b)]%
        {ISP2023}
\bibfield{author}{\bibinfo{person}{Ajay Jaiswal}, \bibinfo{person}{Shiwei Liu}, \bibinfo{person}{Tianlong Chen}, \bibinfo{person}{Ying Ding}, {and} \bibinfo{person}{Zhangyang Wang}.} \bibinfo{year}{2023}\natexlab{b}.
\newblock \showarticletitle{Instant soup: cheap pruning ensembles in a single pass can draw lottery tickets from large models}. In \bibinfo{booktitle}{\emph{Proceedings of the 40th International Conference on Machine Learning}} (<conf-loc>, <city>Honolulu</city>, <state>Hawaii</state>, <country>USA</country>, </conf-loc>) \emph{(\bibinfo{series}{ICML'23})}. \bibinfo{publisher}{JMLR.org}, Article \bibinfo{articleno}{599}, \bibinfo{numpages}{11}~pages.
\newblock


\bibitem[Jaiswal et~al\mbox{.}(2023a)]%
        {EssentialSparsity2023}
\bibfield{author}{\bibinfo{person}{Ajay~Kumar Jaiswal}, \bibinfo{person}{Shiwei Liu}, \bibinfo{person}{Tianlong Chen}, {and} \bibinfo{person}{Zhangyang Wang}.} \bibinfo{year}{2023}\natexlab{a}.
\newblock \showarticletitle{The Emergence of Essential Sparsity in Large Pre-trained Models: The Weights that Matter}. In \bibinfo{booktitle}{\emph{Thirty-seventh Conference on Neural Information Processing Systems}}.
\newblock
\urldef\tempurl%
\url{https://openreview.net/forum?id=bU9hwbsVcy}
\showURL{%
\tempurl}


\bibitem[Rao et~al\mbox{.}(2021)]%
        {DynamicViT2021}
\bibfield{author}{\bibinfo{person}{Yongming Rao}, \bibinfo{person}{Wenliang Zhao}, \bibinfo{person}{Benlin Liu}, \bibinfo{person}{Jiwen Lu}, \bibinfo{person}{Jie Zhou}, {and} \bibinfo{person}{Cho-Jui Hsieh}.} \bibinfo{year}{2021}\natexlab{}.
\newblock \showarticletitle{DynamicViT: Efficient Vision Transformers with Dynamic Token Sparsification}. In \bibinfo{booktitle}{\emph{Advances in Neural Information Processing Systems}}.
\newblock
\urldef\tempurl%
\url{https://arxiv.org/abs/2106.02034}
\showURL{%
\tempurl}


\bibitem[Shen et~al\mbox{.}(2023)]%
        {DataLevelLottery2023}
\bibfield{author}{\bibinfo{person}{Xuan Shen}, \bibinfo{person}{Zhenglun Kong}, \bibinfo{person}{Minghai Qin}, \bibinfo{person}{Peiyan Dong}, \bibinfo{person}{Geng Yuan}, \bibinfo{person}{Xin Meng}, \bibinfo{person}{Hao Tang}, \bibinfo{person}{Xiaolong Ma}, {and} \bibinfo{person}{Yanzhi Wang}.} \bibinfo{year}{2023}\natexlab{}.
\newblock \showarticletitle{Data Level Lottery Ticket Hypothesis for Vision Transformers}. In \bibinfo{booktitle}{\emph{Proceedings of the Thirty-Second International Joint Conference on Artificial Intelligence}} \emph{(\bibinfo{series}{IJCAI-2023})}. \bibinfo{publisher}{International Joint Conferences on Artificial Intelligence Organization}.
\newblock
\urldef\tempurl%
\url{https://doi.org/10.24963/ijcai.2023/153}
\showDOI{\tempurl}


\bibitem[Vaswani et~al\mbox{.}(2017)]%
        {Vaswani2017}
\bibfield{author}{\bibinfo{person}{Ashish Vaswani}, \bibinfo{person}{Noam~M. Shazeer}, \bibinfo{person}{Niki Parmar}, \bibinfo{person}{Jakob Uszkoreit}, \bibinfo{person}{Llion Jones}, \bibinfo{person}{Aidan~N. Gomez}, \bibinfo{person}{Lukasz Kaiser}, {and} \bibinfo{person}{Illia Polosukhin}.} \bibinfo{year}{2017}\natexlab{}.
\newblock \showarticletitle{Attention is All you Need}. In \bibinfo{booktitle}{\emph{Neural Information Processing Systems}}.
\newblock
\urldef\tempurl%
\url{https://api.semanticscholar.org/CorpusID:13756489}
\showURL{%
\tempurl}


\end{thebibliography}

% %%
% %% If your work has an appendix, this is the place to put it.
% \appendix




\end{document}